\documentclass[10pt,journal,compsoc]{IEEEtran}

\usepackage{framed,multirow}

\usepackage[pdftex]{graphicx}

\usepackage{amssymb}
\usepackage{amsfonts}
\usepackage{amsmath}
\usepackage{latexsym}

\usepackage{url}
\usepackage{xcolor}

\begin{document}


\title{Stereo camera system calibration: the need of two sets of parameters}

\author{Riccardo Beschi$^1$, Xiao Feng$^2$, Stefania Melillo$^1$, Leonardo Parisi$^1$, Lorena Postiglione$^1$
        
    \IEEEcompsocitemizethanks{
        \IEEEcompsocthanksitem $^1$ CNR--ISC (National Research Council - Institute for Complex Systems) UOS Sapienza, Rome, Italy 
    }
    
        \IEEEcompsocitemizethanks{
        \IEEEcompsocthanksitem $^2$ College of Engineering, South China Agricultural University 
    }
    
}

\IEEEtitleabstractindextext{

\begin{abstract} 
The reconstruction of a scene via a stereo-camera system is a two-steps process, where at first images from different cameras are matched to identify the set of point-to-point correspondences that then will actually be reconstructed in the three dimensional \textit{real} world. The performance of the system strongly relies of the calibration procedure, which has to be carefully designed to guarantee optimal results. We implemented three different calibration methods and we compared their performance over $19$ datasets. We present the experimental evidence that, due to the image noise, a single set of parameters is not sufficient to achieve high accuracy in the identification of the correspondences and in the $3D$ reconstruction at the same time. We propose to calibrate the system twice to estimate two different sets of parameters: the one obtained by minimizing the reprojection error that will be used when dealing with quantities defined in the $2D$ space of the cameras, and the one obtained by minimizing the reconstruction error that will be used when dealing with quantities defined in the \textit{real} $3D$ world.
\end{abstract}

}

\maketitle

\IEEEdisplaynontitleabstractindextext
\IEEEpeerreviewmaketitle



\section{Introduction}
Stereo camera systems represent an effective instrument for $3D$ reconstruction in great variety of applications and in different fields of science, entertainment, industry and automotive, \cite{entertainment2000, cavagna2017Swarm,dell2014automated, robotics2017,entertainment2007,navigation2019,medicine2016, industrial2020}. 

The common objective of all these applications is the reconstruction of a scene - the one in the common field of view of the cameras - starting from the images. 
The central topic of all stereo-vision applications is then the calibration of the system, which defines the relation between the \textit{real} $3D$ world and the $2D$ world of the cameras. 

The role of the calibration parameters in the reconstruction of the scene is two-fold. On one side they are needed to match the images across the cameras, hence to pair the points belonging to different cameras, images of the same $3D$ target. On the other side they are needed in the actual $3D$ reconstruction process, to define the geometry of the system with respect to the $3D$ reference frame. The calibration procedure has then to be robust against noise in the identification of the correct correspondences, and to be accurate in the $3D$ reconstruction.

The literature suggests different approaches to the calibration problem, as in \cite{2DmeasureCalibration2019a,2DmeasureCalibration2019b, 2DmeasureCalibration2018, 2DmeasureCalibration2016,3DmeasureCalibration2020,3DmeasureCalibration2019,3DmeasureCalibration2016}. 
Regardless the particular strategy, the set of parameters estimated with the calibration procedure is generally used both for the identification of the correct correspondences and for the $3D$ reconstruction. Working with one set of parameters is reasonable in a noise-free world, but it is not sufficient to guarantee optimal results in the \textit{real} world, where images are affected by noise.

We show this discrepancy by implementing three different calibration algorithms. In the first one, i.e. \textit{the essential method}, we calibrate the system estimating the essential matrix from a set of known point-to-point correspondences, following the basic calibration algorithms found in \cite{hartley2003multiple} and in \cite{faugeras}. In the other two methods, we use the set of parameters found with the essential method as the initial condition of a Montecarlo algorithm, to find the parameters that minimize the reprojection error, in what we call the \textit{$2D$ minimization method}, and the reconstruction error, in the \textit{$3D$ minimization method}.

We tested the three methods on $19$ datasets and we show that none of them produces high quality results on the identification of the correspondences and on the $3D$ reconstruction at the same time. The $2D$ minimization algorithm gives the best performance in terms of the identification of the correspondences, while the $3D$ reconstruction algorithm gives the best performance in terms of the accuracy in the $3D$ reconstruction. 

We propose to estimate two different sets of parameters and to use one or the other according to the needs: to identify the correspondences we use the parameters estimated through the $2D$ minimization method, while to reconstruct the already identified correspondences in the $3D$ space we use the  parameters estimated through the $3D$ minimization method.

\section{Background}
In this section we summarize the notations and the basic concepts that we will use throughout the paper. 

\begin{figure}[h]
    \centering
    \includegraphics[width=1.0\linewidth]{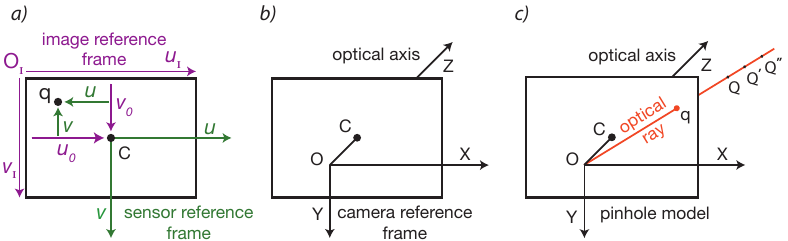}
    \caption{\textbf{The camera.} a. The image and the sensor reference frame. The image reference frame, $O_Iu_Iv_I$ denoted in purple, has the origin on the top left edge of the image, the $u_I$-axis pointing to the right and the $v_I$-axis pointing down. The sensor reference frame, $Cuv$ denoted in green, has the origin in the image center, $C$, the $u$-axis and the $v$-axis parallel to the axes of the image reference frame. b. The camera reference frame, $Oxyz$, has the origin in the camera optical center, the $z$-axis directed as the optical axis, the $x$-axis and the $y$-axis parallel to the $u$-axis and the $y$-axis of the image reference frame. c. the camera pinhole model. The image $q$ of a $3D$ point $Q$ lies at the intersection between the plane of the sensor and the optical ray, denoted in red, connecting $Q$ and the optical center. The camera is then described as a projection centered in the optical center. The relation between a $3D$ point and its image $q$ is not one-to-one: all the $3D$ points lying on the same optical ray passing by $Q$ corresponds to the same image $q$.}
\label{fig::referenceFrame}
\end{figure}

\subsection{Single camera}
\textbf{Pinhole model.} This is the basic camera model adopted in the field of computer vision. It assumes that the $2D$ image, $q$, of a $3D$ target $Q$ lies at the intersection between the camera sensor and the line connecting the camera optical center and the target $Q$, i.e. the optical ray passing through $q$ and $Q$, see Fig.\ref{fig::referenceFrame}c. The target $Q$ and its image $q$ are therefore related by a projection centered on the camera optical center, defined by the following equation:
\begin{equation}\label{eq::qPQ}
    \mathbf{q} = P\mathbf{Q}
\end{equation}
where $\mathbf{q}$ represents the projective coordinates of the $2D$ point $q\equiv(u,v)$, i.e. the three dimensional point $\mathbf{q}\equiv(\bar{u}, \bar{v}, \bar{w})$ such that $u = \bar{u}/\bar{w}$ and $v = \bar{v}/\bar{w}$, $\mathbf{Q}$ represents the projective coordinates of $Q\equiv(X,Y,Z)$, i.e. the four dimensional point $\mathbf{Q}\equiv(\bar{X}, \bar{Y}, \bar{Z}, \bar{W})$ such that $X = \bar{X}/\bar{W}$, $Y = \bar{Y}/\bar{W}$ and $Z = \bar{Z}/\bar{W}$. The $2D$ coordinates $u$ and $v$ of the point $q$ are expressed in the camera reference frame shown in Fig.\ref{fig::referenceFrame}. $P$ is the $3\times 4$ \textit{projective} matrix that defines the relation between the $3D$ world, where coordinates are expressed in meters, and the $2D$ world of the cameras, where coordinates are expressed in pixels. 

\textbf{Projection matrix.} The general form of the projective matrix is the following:
\begin{equation}
    P = K[R|T]
\end{equation}
where $K$ is the $3\times 3$ matrix of the camera internal parameters and $[R|T]$ is the $3\times 4$ matrix of the camera external parameters.

\textbf{Camera internal parameters.}
$K$ is the $3\times 3$ matrix of the camera \textit{internal parameters}, which stores the camera intrinsic characteristics:
\begin{equation}
    K=\left(
    \begin{array}{ccc}
         \Omega & s & u_0  \\
         0 & \Omega & v_0  \\
         0 & 0 & 1
    \end{array}
    \right)
\end{equation}
where $\Omega$ is the focal length expressed in pixel, $s$ is the pixel skewness, $u_0$ and $v_0$ are the coordinate of the image center, $C=(u_0, v_0)$, expressed in the image reference frame, see Fig.\ref{fig::referenceFrame}.

A camera with known internal parameters is referred as \textit{calibrated camera}, and the coordinates of its images may be naturally normalized with the following transformation:
\begin{equation}\label{eq::normalization}
    \hat{q} = K^{-1}q
\end{equation}
where $\hat{q}$ represents the normalized and dimensionless coordinates and  $K^{-1}$ the inverse of the intrinsic parameters matrix.

\textbf{Camera external parameters.}
$[R|T]$ is the $3\times 4$ matrix of the camera \textit{external parameters}, which defines the orientation and position of the camera with respect to the world reference frame, namely $R$ is the $3\times 3$ rotational matrix and $T$ is the three dimensional translational vector that bring the camera reference frame into the world reference frame, i.e. the three dimensional reference frame where the coordinates of the $3D$ point $Q$ are defined. 

\begin{figure}[h]
    \centering
    \includegraphics[width=1.0\linewidth]{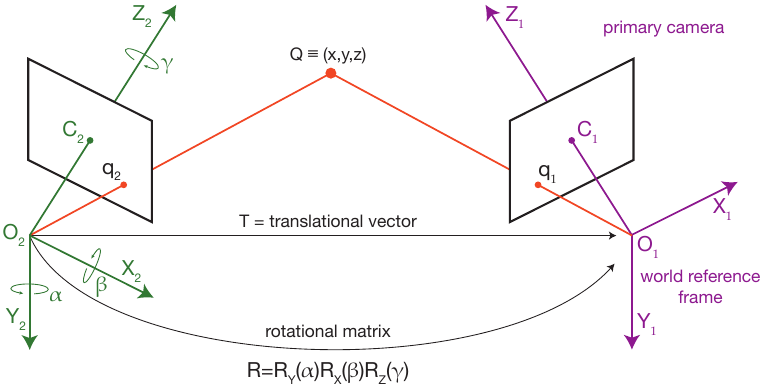}
    \caption{\textbf{Canonical camera system.} The world reference frame coincides with the reference frame of the primary camera, denoted in purple. The primary camera projection matrix is of the form $P_1=K_1[I|0]$ while the projection matrix of the secondary camera is of the form $P_2=K_2[R|T]$, with $K_1$ and $K_2$ represent the internal parameters matrices of the two cameras. $R$ and $T$ are the rotational matrix and the translational three-dimensional vector, which bring the secondary camera reference frame into the world camera frame. By matching the information between the two cameras, the ambiguity of a single camera on the $3D$ reconstruction is overcome: the point $Q$ lies at the intersection of the two optical rays, $r_1$ and $r_2$ denoted in red.}
    \label{fig::worldReferenceFrame}
\end{figure}{}

\begin{figure*}[h]
    \centering
    \includegraphics[width=0.90\linewidth]{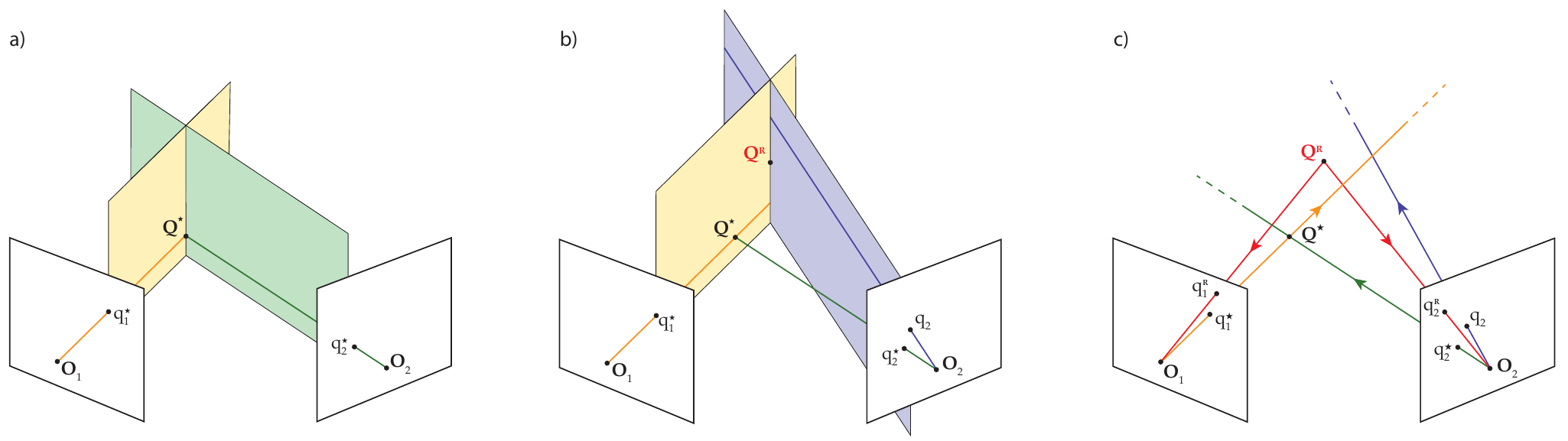}
    \caption{\textbf{The role of noise.} a. In the absence of noise the target $Q^\star$ is imaged in $q_1^\star$ and $q_2^\star$, and it lies at the intersection between the two optical rays, $r_1^\star$ and $r_2^\star$ highlighted in green and orange. b. The noise on the right image produces a mis-positioning of $q_1^\star$ into $q_1$. The two optical rays $r_1$, passing by $q_1$ and highlighted in purple, and $r_2^\star$ do not intersect in the $3D$ space and in principle we cannot reconstruct the correspondence $(q_1, q_2^\star)$ in the $3D$ space. But we can find an approximate solution of the system. The $3D$ point reconstructed from the correspondence $(q_1, q_2^\star)$ is then $Q^R$: the point at the minimum distance from both $r_1$ and $r_2^\star$. c. The reconstructed point $Q^R$ is back-projected , through the two optical lines highlighted in red, onto the two points $q_1^R$ and $q_2^R$, which differ from both the noise-free images, $q_1^\star$ and $q_2^\star$, and from the corrupted image $q_1$. Arrows on the optical lines denote the direction: for the reconstruction we go from the cameras to the $3D$ space (orange, green and purple lines), while for the back-projection we go from the $3D$ space to the cameras (red lines).
}
\label{fig::Noise}
\end{figure*}

\textbf{$\mathbf{3D}$ reconstruction ambiguity.}
The correspondence between the $3D$ world and the $2D$ world of the camera, defined by the central projection in eq.\ref{eq::qPQ}, is not one-to-one: all the points lying on the same optical ray are imaged on the same point, as shown in Fig.\ref{fig::referenceFrame}.
Therefore it is not possible to reconstruct the position of an object using the information of one camera only.

\subsection{Two cameras systems}
The reconstruction ambiguity of the single camera may be overcome matching the information from two cameras. When a $3D$ target $Q$ is seen by two cameras, its two images, $q_1$ and $q_2$, identify two optical rays, $r_1$ and $r_2$ (the two red lines in Fig.\ref{fig::worldReferenceFrame}), at the intersection of which $Q$ lies. The coordinates of $Q$ may be found by solving the following system:
\begin{equation}\label{eq::sysReco}
    \left\{
    \begin{array}{lcl}
         \mathbf{q_1} & = &  P_1\mathbf{Q}\\
         \mathbf{q_2} & = &  P_2\mathbf{Q}\\
    \end{array}
    \right.
\end{equation}
where $\mathbf{q_1}$ and $\mathbf{q_2}$ are the projective coordinates of the two images, $P_1$ and $P_2$ are the projective matrices of the two cameras, and $\mathbf{Q}$ represents the projective coordinates of $Q$.

\textbf{Canonical systems.} In the particular situation where the the world reference frame coincides with the reference frame of one of the two cameras, the system is in its \textit{canonical form}. We denote the reference camera as the \textit{primary} camera, highlighted in purple in Fig.\ref{fig::worldReferenceFrame}, while we denote the other camera as the \textit{secondary} one, highlighted in green in Fig.\ref{fig::worldReferenceFrame}. The projective matrix of the primary camera is of the form $P_1=K_1[I|\mathbf{0}]$, while for the secondary camera the projective matrix is in the standard  form $P_2=K_2[R|T]$, where $K_1$ and $K_2$ are the two internal parameters matrices. 

Note that in this configuration the matrix $R$ and the vector $T$ defines the orientation and position of the secondary camera with respect to the primary camera, hence $[R|T]$ describes the mutual orientation/position of the two cameras, see Fig.\ref{fig::worldReferenceFrame}.

\section{The role of noise}

The pinhole camera model and its generalization to multi-camera system are based on the strong assumption of the absence of noise. This is reasonable for a model but not realistic, since images are in fact affected by noise. The direct effect of noise is a mis-position of the images on the cameras sensors that indirectly affects both the accuracy on the $3D$ reconstruction and the accuracy on the $2D$ reprojection.

Because of the noise, the target $Q^\star$ that should be imaged in the two points $q_1^\star\equiv(u_1^\star, v_1^\star)$ and $q_2^\star\equiv(u_2^\star, v_2^\star)$ is instead imaged in $q_1\equiv(u_1,v_1)$ and $q_2\equiv(u_2,v_2)$, see Fig.\ref{fig::Noise} where for the sake of simplicity we describe the effect of noise on one camera only. As a consequence, the two optical rays $r_1$ and $r_2$ passing by $q_1$ and $q_2$ do not coincide with the two noise-free optical rays, $r_1^\star$ and $r_2^\star$ (passing by $q_1^\star$ and $q_2^\star$). Because of this, $r_1$ and $r_2$ do not intersect in the $3D$ point $Q^\star$, which lies instead at the intersection between $r_1^\star$ and $r_2^\star$, and actually in the general case they do not intersect at all.

Therefore the system defined in eq.\ref{eq::qPQ} does not have an exact solution and in principle we cannot reconstruct the correspondence $(q_1, q_2)$ in the $3D$ space. But we can find an approximate solution of the system: the $3D$ point $Q^R$, defined as the point at the minimum distance from both $r_1$ and $r_2$. This reconstructed $Q^R$ is back-projected onto the two points $q_1^R$ and $q_2^R$, which differ from both the noise-free pair $(q_1^\star, q_2^\star)$, and from the corrupted pair $(q_1, q_2)$, as shown in Fig.\ref{fig::Noise}.

\section{Canonical systems calibration}\label{sec::methods}

We focus here on the calibration of a system in its canonical form. Therefore we assume that the internal parameters of the cameras are known or pre-calibrated and that only the external parameters of the system have to be calibrated.

We implemented three different calibration algorithms: i. the essential method, in which the essential matrix of the system is computed from a set of known point-to-point correspondences; ii. the $2D$ minimization method, which finds the set of parameters that minimizes the reprojection error over the same set of known point-to-point correspondences already used in the essential method; iii. the $3D$ minimization method, which finds the set of parameters that minimizes the reconstruction error over a set of target-to-target and point-to-point correspondences.

\subsection{Essential method.} 
Following \cite{hartley2003multiple} and \cite{faugeras}, we start from a set of $N\geq 5$ point-to-point correspondences within the two cameras defined as 
\begin{equation}\label{eq::Q2D}
    \mathbb{Q}_{2D}=\{q_{1k}, q_{2k}\}_{\{k=1,\cdots,N\}}
\end{equation} 
where $q_{1k}$ and $q_{2k}$ are the images of the same target in the primary and secondary camera respectively.

We use this set to estimate the essential matrix $E$, namely the $3\times 3$ matrix that minimizes the sum of the \textit{residuals}, $R_k$:
\begin{equation}\label{eq::residuals}
    R_k = \mathbf{\hat{q}}_{1k}^\mathrm{T}E\mathbf{\hat{q}}_{2k}
\end{equation}
where $k=1,\cdots,N$, while $\mathbf{\hat{q}}_{1k}$ and $\mathbf{\hat{q}}_{2k}$ represent the projective coordinates of $q_{1k}$ and $q_{2k}$ normalized as in eq.\ref{eq::normalization}\footnote{Two constraints are also added to the minimization of the residuals: i. $det E = 0$; ii. $E$ has only one not null eigenvalue with multiplicity equal to two.} and the superscript $\mathrm{T}$ denotes the transpose of the vector.

The essential matrix is of the form:
\begin{equation}\label{eq::E}
    E = [t]_xR
\end{equation}
where $R$ is the rotational matrix and $t$ the unitary vector associated to translation vector $T$ ($t=T/|T|$) that bring the reference frame of the secondary camera into the primary camera, and $[t]_x$ denotes the cross product expression for the vector $t$\footnote{
\begin{equation*}
    [t]_x = \left(
    \begin{array}{ccc}
         0 & -t_z & t_y\\
         t_z& 0 & -t_x\\
         -t_y & t_x & 0\\
    \end{array}
    \right)
\end{equation*}
where $t_x$, $t_y$ and $t_z$ denote the $t$ components along the $x$-axis, the $y$-axis and the $z$-axis respectively.
}.

We express $t$ in terms of the polar and zenith angles associated to the center of the primary reference frame in the secondary camera reference frame:
\begin{equation}
    t = (\cos\delta\cos\varepsilon, \cos\delta\sin\varepsilon, \sin\delta)
\end{equation}
and
$R$ in terms of the three angles of yaw ($\alpha$), pitch ($\beta$) and roll ($\gamma$) of the secondary camera with respect to the primary camera reference frame:
\begin{equation}
    R = R_y(\alpha)\cdot R_x(\beta)\cdot R_z(\gamma)
\end{equation}
where $R_x$, $R_y$ and $R_z$ represent the rotations about the $x$-axis, the $y$-axis and the $z$-axis respectively. 
The essential matrix has then $5$ degrees of freedom, represented by the five angles: $\alpha$, $\beta$, $\gamma$, $\delta$ and $\varepsilon$. 

Inverting eq.\ref{eq::E} we find $R$ and $t$ from the estimated essential matrix $E$. Hence we can compute the $[R|T]$ matrix up to the scale factor $|T|$, which represent the distance between the centers of the two camera reference frames, as shown in Fig.\ref{fig::worldReferenceFrame}. We fix this scale factor measuring the system baseline, $|T|$, in the \textit{real} $3D$ world.

\subsection{2D and 3D minimization methods.} Both these methods use the parameters estimated through the essential matrix method and the associated angles, $\alpha$, $\beta$, $\gamma$, $\delta$ and $\varepsilon$, as the initial condition of a iterative Montecarlo procedure.

In detail, at each iteration $i$:

\noindent
\textbf{Step 1.} We randomly select one of the five angles. 

\noindent
\textbf{Step 2.} We randomly choose if to add or subtract a quantity $\Delta$ (initially set equal to $0.001$rad) to the angle selected at Step 1. 

\noindent
\textbf{Step 3.} We define the set of angle for the current iteration, $(\alpha_i, \beta_i, \gamma_i, \delta_i, \varepsilon_i)$, obtained from the angles of the previous iteration modifying the angle selected in Step 1 of the quantity $\pm\Delta$.

\noindent
\textbf{Step 4.} We compute the rotation matrix and the translation vector that correspond to $i$-th set of angles: $R_i=R_y(\alpha_i)R_x(\beta_i)R_z(\gamma_i)$ and $T_i=|T| (\cos\delta_i \cos\varepsilon_i, \sin\delta_i \cos\varepsilon_i, \sin\varepsilon_i)$.

\noindent
\textbf{Step 5.} We compute a cost function, defined in detail in Section\ref{sec::costFunction}, $C_i=C(R_i,T_i)$ over a set of known correspondences (point-to-point correspondences for the $2D$ minimization method and target-to-point correspondences for the $3D$ minimization method) and we compare this current cost with the one of the previous iteration, $C_{i-1}$. 

\noindent
If $C_i<C_{i-1}$ we accept the move. 

\noindent
Otherwise we do not accept the move, hence we set the current set of angles to the ones of the previous iteration: $(\alpha_i, \beta_i, \gamma_i, \delta_i, \varepsilon_i)=(\alpha_{i-1}, \beta_{i-1}, \gamma_{i-1}, \delta_{i-1}, \varepsilon_{i-1})$.

\noindent
\textbf{Step 6.} We compute the acceptance ratio, i.e. the ratio between the accepted moves and the total number of iteration. If the acceptance ratio is smaller than a chosen value ($0.2$ in our case), we reset the number of iteration to $0$ and we decrease $\Delta$ (in our case we multiply $\Delta$ by a factor equal to $0.75$). If $\Delta$ is larger than a threshold $\Delta_{min}$ ($10^{-6}$rad in our case) we go back to Step1, otherwise the procedure ends and the solution of the minimization problem is the pair $(R_i, T_i)$.

\subsubsection{Cost functions}\label{sec::costFunction} 
The $2D$ minimization method and the $3D$ minimization method differ for the cost functions used. 

\textbf{$\mathbf{2D}$ minimization method.} Given the set $\mathbb{Q}_{2D}$ of point-to-point correspondences defined in eq.\ref{eq::Q2D}, we first reconstruct each pair in the $3D$ space and then we back-project these $3D$ reconstructed points onto the $2D$ planes of the camera sensors, obtaining the set of reprojected pairs $\mathbb{Q}_{2D}^R=\{(q_{1k}^R, q_{2k}^R)\}_{\{k=1,\cdots,N\}}$. 

We define the reprojection error as the $2D$ distance between the original position $q_{jk}$ and its reprojection $q_{jk}^R$:
\begin{equation}\label{eq::reproError}
    E_{2D}(j,k) = |q_{jk}-q_{jk}^{R}| \quad\mbox{with} j=1,2 \mbox{ and } k=1,\cdots,N
\end{equation}

Finally we define the reprojection cost as the sum of the reprojection errors over all the points $q_{jk}$.

\textbf{$\mathbf{3D}$ minimization method.} We start from a set of pairs of $3D$ targets, $\mathbb{Q}_{jk}=\{Q_j, Q_k\}$, at the mutual distances $D_{jk}=|Q_j-Q_k|$ that is measured in the \textit{real} world. We define the set $\mathbb{Q}_{3D}$ as:
\begin{equation}\label{eq::recoError}
    \mathbb{Q}_{3D}=\{D_{jk}, (q_{1j}, q_{2j}), (q_{1k}, q_{2k})\}_{\{j,k=1\cdots,N\}}
\end{equation} 
where $(q_{1j}, q_{2j})$ and $(q_{1k}, q_{2k})$ are the images of $Q_j$ and $Q_k$ on the two cameras.

From the pairs $(q_{1j}, q_{2j})$ and $(q_{1k}, q_{2k})$ we obtain $Q_j^R$ and $Q_k^R$, the reconstructed positions of the two targets $Q_j$ and $Q_k$, and we compute their distance $D_{jk}^R=|Q_j^R-Q_k^R|$.

We define the $3D$ reconstruction error between the $j$-th and the $k$-th targets as the difference between their measured distance and their reconstructed distance:\footnote{The same procedure may be applied at the set of $3D$-$2D$ correspondences defined as $\mathbb{Q}_{3D}=\{Q_k, (q_{1k}, q_{2k})\}$ made of $3D$ targets and their two images. The reconstruction error should then be defined as $E_{3D}(k)=|Q_k-Q_k^R|$ where $Q_k^R$ is the $3D$ point reconstructed from $q_{1k}$ and $q_{2k}$. But in this approach we would need to accurately measure the position of the target $Q_k$ in the reference frame of the primary camera, which is not a physical reference frame. We chose to measure a quantity that does not depend on the particular reference frame, but that guarantees that the metric proportion of the scene are respected and also the dynamic quantities of the targets, i.e. velocity and acceleration, may be accurately computed.}:
\begin{equation}
    E_{3D}(j,k) = |D_{jk}-D_{jk}^{R}|
\end{equation}

Finally, we define the reconstruction cost as the sum of the reconstruction errors over all the elements of $\mathbb{Q}_{3D}$.

\begin{figure*}[h]
    \centering
    \includegraphics[width=0.86\linewidth]{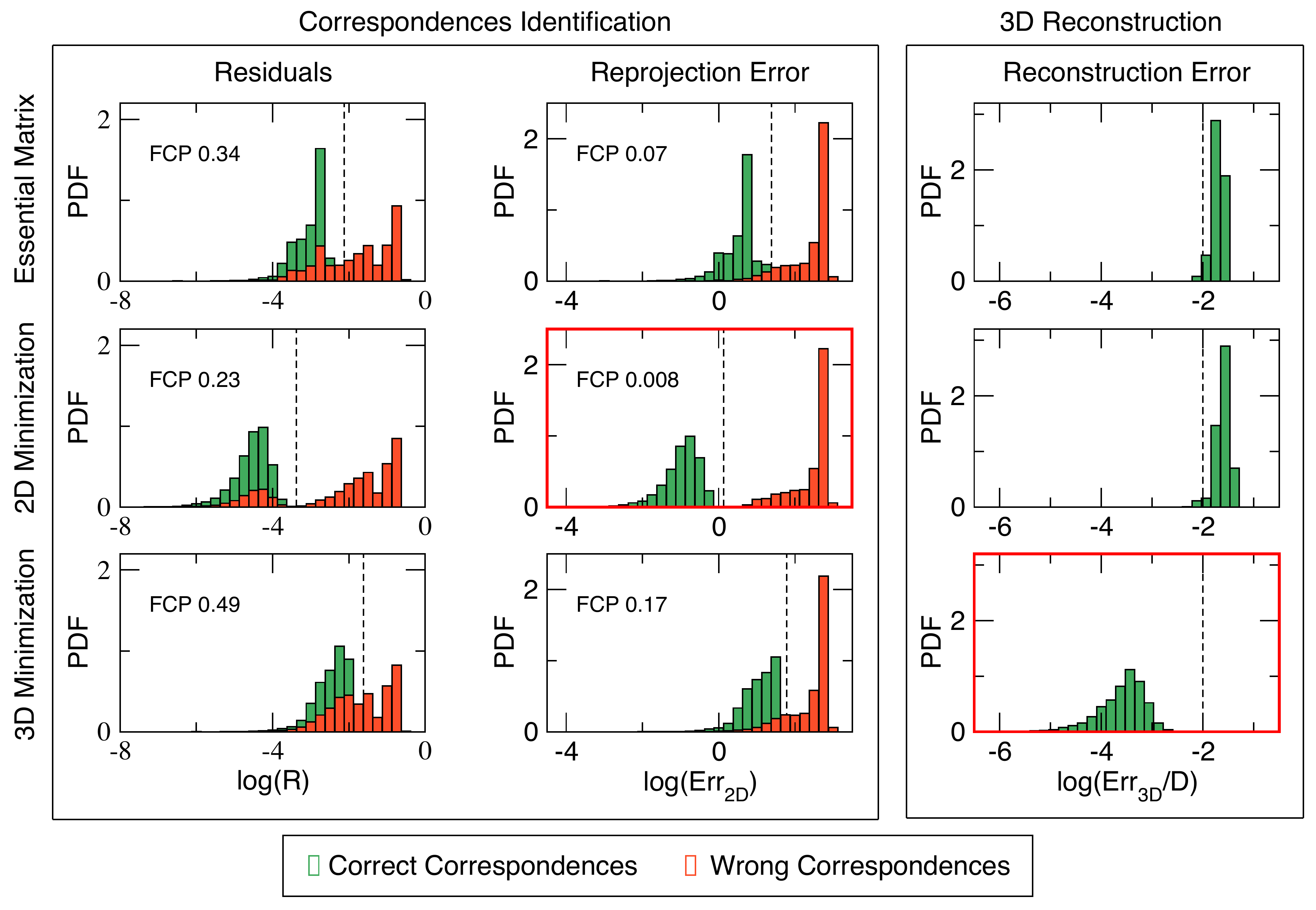}
    \caption{\textbf{Results.} The comparison between the three calibration methods we implemented: the essential method (first row), the $2D$ minimization method (second row) and the $3D$ minimization method (third row). Correspondences identification (first two columns): it is evaluated by computing the probability distribution functions of the log of the residuals (first column) and of the log of the $2D$ reprojection error (second column) on the set of correct and wrong correspondences (highlighted in green and orange respectively). The quality of the method and on the quantity used to discriminate the correct correspondences from the wrong ones is quantified as the FCP (false correspondence probability), obtained computing the overlapping area between the correct and the wrong PDFs. The dashed black lines represent the right tail of the correct correspondence PDF that is not always visible. $3D$ reconstruction (third column): it is evaluated by computing the PDF of the log of the percentage $3D$ reconstruction error. The black dashed lines at $-2$ (equal to a percentage error of $0.01$) represent the acceptable threshold in many applications. The two plots with the red frames show that the best performance in terms of the identification of the correspondences is achieved using the $2D$ minimization method and using the reprojection error, while the best performance in terms of $3D$ reconstruction is achieved using the $3D$ minimization method.}
\label{fig::results}
\end{figure*}

\section{Datasets}\label{sec::datasets}
We tested the calibration accuracy on $19$ sets of images acquired with a system of two cameras (IDT-M5) equipped with $50$mm lenses (Schneider Xenoplan $2.8/50$). We calibrate the internal parameters of each camera separately in the lab, using a method based on \cite{StandardCalib}: we collect $50$ images of a $13\times 19$ checkerboard in different positions, we randomly pick $20$ of these pictures and we estimate the focal length, the position of the image center and the distortion coefficients. We iterate this process $50$ times and we choose each parameter as the median value obtained in the iterations.

The datasets used in the tests are sets of $25$ images of two calibration targets, namely two $2\times 2$ checkerboards mounted on an aluminum bar, whose centers will be denoted by target $A$ and $B$ in the following. We detect the position of the targets on the images with the automatic subpixel routine in \cite{subPixel}.

The datasets slightly differ in the cameras baseline, $|T|\in[3.449m, 5.936m]$, and in the distance between the two targets, $D\in[0.811m, 1.027m]$, which is constant within the images of the same dataset. We accurately measure $|T|$ and $D$ of each dataset with a laser range finder.

For each dataset we define the set $\mathbb{Q}_{2D}$ of point-to-point correspondences as:
\begin{equation}
    \mathbb{Q}_{2D} = \{(q_{1k}^A, q_{2k}^A)\cup(q_{1k}^B, q_{2k}^B)\}_{\{k=1,\cdots,25\}}
\end{equation}
where $(q_{1k}^A, q_{2k}^A)$ and $(q_{1k}^B, q_{2k}^B)$ are the images on the $k$-th picture of the target $A$ and $B$ respectively. 

We also defined the set $\mathbb{Q}_{3D}$ of the $3D-2D$ correspondences as:
\begin{equation}
    \mathbb{Q}_{3D}=\{D, (q_{1k}^A, q_{2k}^A), (q_{1k}^B, q_{2k}^B)\}_{\{k=1,\cdots,25\}}
\end{equation}

\section{Tests and results}\label{sec::tests}
We tested the three algorithms described in Section \ref{sec::methods} on the datasets described in Section \ref{sec::datasets}. We used $20$ images, i.e. calibration images, of the total $25$ images of each dataset to estimate the system parameters and the other $5$ images, i.e. validation images, to validate the calibration. To increase the statistical significance of the tests, for each dataset we performed $100$ runs, choosing at each run a different set of calibration images.

In each run we performed the three algorithms described above, obtaining three different sets of parameters of the system: i. from the essential method we estimated the essential matrix, $E$, and inverting eq.\ref{eq::E} we found its associated $[R|T]$ matrix; ii. from the $2D$ minimization method we estimated the $[R|T]_{2D}$ matrix, and we computed the associated essential matrix, $E_{2D}$, from eq.\ref{eq::E}; iii. from the $3D$ minimization method we estimated the $[R|T]_{3D}$ matrix, and we computed the associated essential matrix, $E_{3D}$, from eq.\ref{eq::E}.

We evaluated the three calibration algorithms in terms of the accuracy in the identification of the point-to-point correspondences across the cameras and in terms of the accuracy in the $3D$ reconstruction. 

\noindent
\textbf{Correspondences identification.} For the identification of the correspondences we need to define a quantity that discriminates the correct correspondences, i.e. the pairs of points images of the same target in the two cameras, from the wrong ones, i.e. pairs of points that do not correspond to the same target. 

We used the $5$ validation images of each run to define the set of correct correspondences:
\begin{equation}
\mathbb{Q}_{2D}^C = \{(q_{1k}^A, q_{2k}^A)\cup(q_{1k}^B, q_{2k}^B)\}_{\{k=1,\cdots,5\}}
\end{equation}

as the set of the pairs of points (one for each camera) both corresponding to the target $A$ or the target $B$ on the $k$-th image.

But we also used the same validation images to define the set of wrong correspondences:
\begin{multline}
    \mathbb{Q}_{2D}^W = \{(q_{1k}^A, q_{2k}^B)\cup(q_{1k}^B, q_{2k}^A)\cup(q_{1j}^A, q_{2k}^B)\cup(q_{1j}^B, q_{2k}^A)\cup
                      \\\cup(q_{1j}^A, q_{2k}^A)\cup(q_{1j}^B, q_{2k}^B)\}_{\{j,k\in[1,\cdots,5]\mbox{ and } j\neq k\}}
\end{multline}
as the set of the pairs obtained associating the point corresponding to target $A$ to the point corresponding to target $B$ on the same image, $(q_{1k}^A, q_{2k}^B)\cup(q_{1k}^B, q_{2k}^A)$, or on different images, $(q_{1j}^A, q_{2k}^B)\cup(q_{1j}^B, q_{2k}^A)$, and of the pairs of points both corresponding to the same target $A$ or $B$ in different images, $(q_{1j}^A, q_{2k}^A)\cup(q_{1j}^B, q_{2k}^B)$.

The two natural quantities for the identification of the correspondences are the residuals, defined in eq.\ref{eq::residuals}, and the reprojection error, defined in eq.\ref{eq::reproError}. We computed the residuals using the three different essential matrices $E$, $E_{2D}$ and $E_{3D}$, and the reprojection errors using the three different matrices of the external parameters, $[R|T]$, $[R|T]_{2D}$ and $[R|T]_{3D}$. Finally we computed the probability distribution functions (PDF) shown in the first two columns of Fig.\ref{fig::results}.

To discriminate the correct correspondences from the wrong ones, the two PDFs should be separated, namely the right tail of the distribution of the correct correspondences (highlighted with a black dashed line on the plots in the first two columns of Fig.\ref{fig::results}) should not overlap the left tail of the distribution of the wrong correspondences. We measured the \textit{false correspondence probability} (FCP) as the overlapping area between the correct and wrong probability distributions, so that an high value of FCP corresponds to a high probability of a mis-correspondence. 

In Fig.\ref{fig::results} we reported the FCP for the three algorithms and for both the residual and the reprojection error. The best performance is obtained when we use the $2D$ calibration method and we identify the correspondences through the reprojection error: the plot with the red frame in the second column of Fig.\ref{fig::results}.

\noindent
\textbf{3D reconstruction.} To evaluate the accuracy in terms of the $3D$ reconstruction we measure the $3D$ reconstruction error defined in eq.\ref{eq::recoError}. In the last column of Fig.\ref{fig::results} we show the probability distribution of the percentage reconstructed error, i.e. $Err_{3D}/D$, for the three algorithms we implemented. We chose to compute and to plot the percentage reconstruction error, because this is the measure that it is relevant in most of the application: an error of $1mm$ over a distance of $1m$ has a different meaning than an error of $1mm$ over a distance of $1cm$. 

The best performance is obtained using the $3D$ calibration method, i.e. the plot with the red frame in the last column of Fig.\ref{fig::results}. The dashed black line at $-2$ corresponds to a percentage error equal to $0.01$, which is an acceptable accuracy for most of the applications. Note that, while the reconstruction error for the $3D$ method is always smaller than the $0.01$, for the other two methods errors are always above this threshold. 

\section{The need of two sets of parameters}
The tests presented in Section \ref{sec::tests} show that using a single set of parameters for both the identification of the correspondences and for the $3D$ reconstruction is not optimal. With the $2D$ minimization algorithm we can successfully detect the correct correspondences computing their $2D$ reprojection error, but we are not accurate in the $3D$ reconstruction. On the opposite, with the $3D$ minimization algorithm we achieve a very high accuracy in the $3D$ reconstruction, but we cannot discriminate the correct correspondences from the wrong ones. 

The reason for this double-set of parameters has to be found in the presence of noise and in how the two methods manage it. 

In the $2D$ minimization method we minimize the reprojection error, hence we are implicitly assuming that the position of the images on the camera sensors are not affected by noise and that all the noise is in the $3D$ space. Given a target $Q$ and its images $q_1$ and $q_2$, we look for those parameters such that the two optical lines, $r_1$ and $r_2$, passing by $q_1$ and $q_2$ intersect in the $3D$ space in a point $Q^R$, the reprojections of which are as close as possible to the original points. It does not matter how far the point $Q^R$ is from the original $3D$ point $Q$, the only significant quantity here is the reprojection error.

On the opposite, in the $3D$ minimization method we minimize the reconstruction error, hence we are implicitly assuming that the position of the $3D$ targets are not affected by noise and that all the noise is in the $2D$ space. Given two targets $Q^A$ and $Q^B$, their distance in the $3D$ space and their images $(q_1^A, q_2^A)$ and $(q_1^B, q_2^B)$, we look for those parameters such that the $3D$ distance between the two reconstructed points is as close as possible to the measured distance. It does not matter if the two pairs of optical lines, $(r_1^A, r_2^A)$ and $(r_1^B, r_2^B)$ passing by the images of the two targets, intersect in the $3D$ space, but only that the distance between the two reconstructed points is close to the reality. The only significant quantity here is the $3D$ distance between the targets: no matter how large the reprojection error is.

\section{Conclusions}
We implemented three different calibration methods for the estimation of the external parameters of a camera system with the aim of comparing their quality in terms of two factors: i. the correct identification of point-to-point correspondences; ii. the accuracy of the $3D$ reconstruction.

We tested the three methods over $19$ datasets and we presented the experimental evidence that, due to the image noise, a single set of parameters cannot be optimal both for the identification of the correspondences and for the $3D$ reconstruction. 

To obtain good results in terms of the identification of the correspondences, we need to work in the $2D$ space of the cameras minimizing the $2D$ reprojection error and moving all the noise in the $3D$ space, hence producing high $3D$ reconstruction error. On the opposite to obtain good results in terms of $3D$ reconstruction, we need to work in the $3D$ space minimizing the $3D$ reconstruction error and moving all the noise on the $2D$ space of the cameras, hence producing large reprojection errors and a high probability of false correspondences.

The optimal choice for the system calibration is then to calibrate the system twice and to use the two sets depending on the quantities that we need to compute. The optimal set of parameters for the computation of all those quantities that live in the $2D$ space of the camera is the one obtained by minimizing the reprojection error, while the optimal set of parameters for the computation of all those quantities that live in the $3D$ \textit{real} world is the one obtained by minimizing the $3D$ reconstruction error.

We do not claim that the three algorithms presented in the paper are an exhaustive panorama of all the calibration techniques provided by the literature. But the same tests that we presented here may be applied to different and even more complicated and more accurate algorithms. The core message of the paper is that for the two tasks the calibration is meant for, we are in the usual \textit{blanket-too-short} dilemma, so that if we improve the accuracy in the $2D$ space we lose accuracy in the $3D$ space and viceversa, hence the easiest and optimal approach is to separate the two tasks and calibrate the system twice.

\section*{Acknowledgments}
This work was supported by ERC grant RG.BIO (Grant No. 785932).

\bibliographystyle{IEEEtran}
\bibliography{IEEEabrv,biblio}



\end{document}